# GuideMe: A Mobile Application based on Global Positioning System and Object Recognition Towards a Smart Tourist Guide

Wadii Boulila
*National School of Computer Sciences*
*University of Manouba*
*Manouba, Tunisia*
*College of Computer Science and Engineering*
*Taibah University*
*Medina, Saudi Arabia*
wadii.boulila@riadi.rnu.tn

Anmar Abuhamdah
*College of Business Administration*
*Taibah University*
*Medina, Saudi Arabia*
aabuhamdah@taibahu.edu.sa

Maha Driss
*National School of Computer Sciences*
*University of Manouba*
*Manouba, Tunisia*
*College of Computer Science and Engineering*
*Taibah University*
*Medina, Saudi Arabia*
maha.idriss@riadi.rnu.tn

Slim Kammoun
*College of Computer Science and Engineering*
*Taibah University*
*Medina, Saudi Arabia*
*LaTICE Laboratory, Tunis* University, Tunisia
kammslim@gmail.com

Jawad Ahmad
*School of Computing, Edinburgh Napier University, United Kingdom*
*Medina, Saudi Arabia*
j.ahmad@napier.ac.uk

*Abstract*—Finding information about tourist places to visit is a challenging problem that people face while visiting different countries. This problem is accentuated when people are coming from different countries, speak different languages, and are from all segments of society. In this context, visitors and pilgrims face important problems to find the appropriate doaas when visiting holy places. In this paper, we propose a mobile application that helps the user find the appropriate doaas for a given holy place in an easy and intuitive manner. Three different options are developed to achieve this goal: 1) manual search, 2) GPS location to identify the holy places and therefore their corresponding doaas, and 3) deep learning (DL) based method to determine the holy place by analyzing an image taken by the visitor. Experiments show good performance of the proposed mobile application in providing the appropriate doaas for visited holy places.

## I. Introduction

Every year, millions of Muslims perform the pilgrimage (Hajj) rituals in Makkah Al Mukarramah (in Saudi Arabia). These pilgrims come from various countries of the world and from all segments of society (whether male or female, adults or children, literate or illiterate). For instance, the number of pilgrims or visitors for the year 2019 reached 2,489,406 visitors, where 74.5% (i.e. 1,855,027) of the visitors representing the visitors that come from outside Saudi Arabia [1] [2]. Generally, the visitors may need to say various phrases or sentences (doaas) while performing the ritual pilgrimage or traveling between different holy places. Many visitors face problems finding the appropriate doaas for a given holy place, Sometimes, some visitors use books to find out the appropriate doaas (which is difficult to use), or ask other people for it (in which may get wrong information).

Currently, several existing mobile applications help in solving some problems and become a solution to help people and visitors such as presented in [3] [4] [5]. However, most of them do not help provide general doaas and could not help determining the doaas according to the current location of visitors. Today, most of the visitor mobiles have a camera and are embedded with a powerful processor that supports mobile applications based on deep learning (DL). The main idea of this paper is to detect holly places present in images and determine the appropriate doaas for them. On the other hand, the Global Positioning System (GPS) also may assist pilgrims by determining their locations and therefore providing the appropriate doaas according to their locations.

Several DL-based approaches are proposed for image processing [6] [7]. However, most of them require a powerful CPU and memory to run efficiently. It has been proven that it is difficult to achieve the required and effective performance when executing DL networks on edge devices. One solution that can be used to overcome these limitations is to use cloud computing or deep learning on the edge. This latter means performing data processing and executing DL networks locally on the user's device without dealing with a third party to process data [8] [9] [10].

The goal of this paper is to develop a solution that supports pilgrims and visitors in finding the appropriate doaas for a given holy place. The proposed solution is based on three options: 1) determining the appropriate doaas by a manual search, 2) using the GPS location to determine the appropriate doaa, and 3) using a DL-based method to identify the holy place and doaas for this place by analyzing an image taken by the visitor. To the best of the authors' knowledge, the proposed application is the first of its kind in the market that determines the appropriate doaas based on an image taken by the visitor and by the GPS location.

The proposed solution consists of two steps as follow:

1) Off-line processing: during this process images of holy places are collected, and a training process is performed to be able to identify them. Then, the trained model is converted to lite format to be deployed on the client devices when downloading the proposed application.

2) On-line processing: the holy place present in the image taken by the visitor is recognized by the model that was deployed on the user's device without the need to connect to the internet.

This paper is outlined as follows. Section II presents the related works. Section III describes the proposed method. Section IV describes experimental results and discussion. Finally, section V presents the conclusion of this paper and future work.

## II. RELATED WORKS

DL is considered as one of the most popular research directions in the current time providing promising results in many fields [11-18]. Recently, various research works have been conducted based on DL techniques to solve the problem of object detection on edge devices. Salonidis et al. [19] proposed DeepCham as a solution for context variations caused by different locations. DeepCham is a DL-based framework that allows object recognition in a mobile device. The proposed mobile application generates high-quality training using a distributed algorithm and a user labeling process by using a new dataset that has different locations and types. The DeepCham depends on the distributing algorithm, which enhances the training of the model. Experiments conducted on DeepCham have shown that the approach is very effective and accurate using a generic deep model.

Liu et al. [20] developed a system for food assessment accuracy. The authors developed a DL-based algorithm that visually recognizes food based on its category and division. Liu et al. designed a food recognition system running on edge to outrun the traditional problems of cloud computing. The proposed system is characterized by high accuracy, fast response time, and reduction of electricity use.

Zeng et al. [21] proposed a pill recognition design and assessment to recognize unknown prescription pills using smartphones. The system, called MobileDeepPill, uses a triplet loss function and a multi-CNNs model to identify the pill characteristics. The system requires 34MB runtime memory to run the model. The performance improved from 1:58s to 0.27s. The pill image recognition may impact millions of people and healthcare society. The system is based on a multi-CNNs architecture, which will increase the accuracy of the model.

Picon et al. [22] developed an algorithm to detect multiple plant diseases. The authors analyze the performance of early disease identification using more than 8178 pictures in two areas. Results show that the accuracy has been improved from 0.87 to 0.96 in Germany.

Ran et al. [23] developed an offloading strategy to enhance DL interaction. The framework links front-end devices with the strongest backend. The developed strategy determines the model accuracy, quality, and constraints. The authors proposed an approach, called Deep-Decision, that can choose the suitable and convenient model. The results show that deep-decision is able to make smart decisions under variable internet, unlike other approaches that neglect video bit rate.

Nikouei et al. [24] presented an approach based on edge computing for real-time human detection. The authors developed a Lightweight Convolutional Neural Network (LCNN) algorithm to detect pedestrians using an edge device, which will save the processing time and the heavy loading of the huge amount of data on networks. A prototype has been implemented on Raspberry PI 3 using OpenCV libraries. The model was trained using ImageNet and VOC07 datasets which contain the objects of interest. Experiments have shown that LCNN has a promising future in applications on the edge.

Tan and Cao [25] described DL-related issues and the adequate solutions for these issues. DL has been used to analyze the videos in some mobile applications. Although DL models provide high-quality results, however, they consume energy when running on devices. The main challenge is to provide max accuracy in a few running times. The authors proposed to use the Neural Processing Unit (NPU) and FastVA, which supports video analysis on the edge.

## III. PROPOSED MOBILE APPLICATION

Nowadays, several digital services are offered to pilgrims and support them in different activities related to Hajj and Umra. In the field of doaa-support applications, there are several shortcomings and a lack of interest in the existing mobile applications. However, the number of pilgrims is increasing every year; therefore, there is a need to provide an application that helps unlettered pilgrims and also pilgrims that are visiting holy places for the first time.

The proposed system is a mobile application, called smart doaa, that provides doaas related to places to visit according to three different options as depicted in Fig. 1:

- Option 1: display the appropriate doaas by identifying the holly place in an image taken by the user.
- Option 2: display the appropriate doaas based on the location of the user.
- Option 3: display the appropriate doaas based on manual search.

These options ensure more flexibility in using the proposed mobile application.

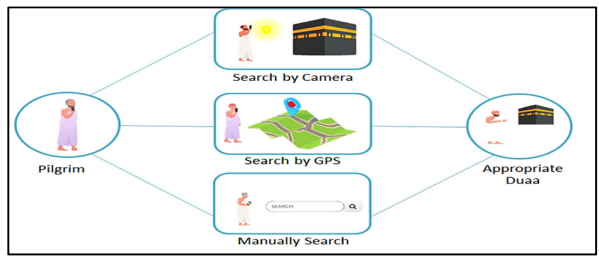

Fig. 1. GuideMe algorithm steps.

The general algorithm of the smart doaas application is presented in Fig. 2. At the beginning of the algorithm, the pilgrim has the choice to select between the three options to display doaas. The view list will display all available doaas in the database then the visitor will select one of them. In the case of selecting object detection, the application requires that the visitor takes a picture of his place. The application will identify if there a holy place in the taken picture based on the trained DL model. If there is a holy place in the picture, then the application will display its corresponding doaas.

```
Algorithm 1 General Smart Dua
 1: input user selection
 2: Database ← ConnectDatabase
 3: Duas[..] ← get all duas from database
 4: if selection = list then                              ▷ To choose manually
 5:     Duas.display
 6:     input : selectionIndex
 7: else if selection = location then         ▷ automatically using the location
 8:     if GPS = disabled then                                  ▷ checking GPS
 9:         GPSactivationRequest
10:     else if permission = ungranted then           ▷ checking permission
11:         requestPermission
12:     userLocation ← currentLocation
13:     places ← get all places from database
14:     for each place in places do                    ▷ Getting place name
15:         if Distance(place.location, currentLocation) < 21 then
                         ▷ Getting place when the distance less than 21 meters
16:             currentPlace ← place
17:             break
18:         end if
19:     end for
20:     for index ← 0 to Duas.size do
21:         if Duas[index].placeId = currentPlace.id then
                                         ▷ Getting Dua with same id as current place
22:             selectionIndex ← index
23:         end if
24:     end for
25: end if
26: else if selection = objectDetection then
                                         ▷ automatically using deep learning
27:     input : Image                                  ▷ Image from camera
28:     Place ← DeepLearningModelProcessing(Image)
                                         ▷ Deep Learning Model Processing algorithm
29:     for index ← 0 to Duas.size do                  ▷ Getting Dua index
30:         if Duas[index].place = Place then
                                         ▷ dua place equal image place
31:             selectionIndex ← index
32:         end if
33:     end for
34: end if
35: dua ← Duas[selectionIndex]                    ▷ Getting dua using the index
36: return dua
```

Fig. 2. Main steps of the proposed general algorithm.

The process of detecting the existence of a holy place in a picture is detailed in Algorithm 2 (see Fig. 3). This algorithm requires an image to process and returns the name of the holy place that the image contains if any. To process the image, the application transforms it into a FirebaseVisionImage. After that, the DL model will be called to work locally by creating an instance of the OnDeviceAutoMLImageLabeler. The result will be an array that contains place names and confidence of each place. The application will display holy places sorted by confidences descending and having a value greater than 0.8.

```
Algorithm 2 processing an image using deep learning model
 1: input Image
 2: VisionImage ← FirebaseVisionImage(Image)
       ▷ Creating a FirebaseVisionImage to pass it to FirebaseVisionImageLabeler
 3: localModel ← FirebaseAutoMLLocalModel("manifest.json")
                         ▷ Specifying the path to the model manifest file
 4: options ← setOnDeviceAutoMLModel(localModel)
                                              ▷ Configure model sources
 5: options ← setConfidenceThreshold(0.5f)
                                              ▷ Determine an appropriate value
 6: labeler ← instanceOfFirebaseVisionImage(options)
                                              ▷ Create an image labeler from the model
 7: labelers[] ← label.processImage(VisionImage)
                                              ▷ Processing the image to get the results
 8: if labelers is not empty then
                                              ▷ Checknig if there's any result to continue
 9:     maxConfidence ← labelers[0].confidence
10:     for each labeler in labelers do
                                              ▷ Walking through all labeler
11:         if maxConfidence < labeler.confidence and labeler.confidence =>
    0.8 then
                                              ▷ Highest similarity
12:             maxConfidence ← labeler.confidence
13:             PlaceName ← labeler.text
14:         end if
15:     end for
16: end if
17: return PlaceName                    ▷ The place name with highest similarity
```

Fig. 3. Deep learning model processing algorithm.

In case of selecting the location-based method, the application will check whether the GPS is enabled and permissions are granted, otherwise, it will request permission. In case of permission granted, the application will get the current location of the visitor and compare it with all stored holy places in the database. This is done using distance calculation computed by Algorithm 3 (see Fig. 4). If the distance is less than 20 meters, then it will display the current holy place and provides its corresponding doaas. Algorithm 3 uses "the 'haversine' formula to calculate the great-circle distance between two points – that is, the shortest distance over the earth's surface – giving an "as-the-crow-flies" distance between the points" [26]. At the end of the algorithm, it returns the distance in meters.

```
Algorithm 3 Calculate distance
 1: Input point1, point2
 2: earthRadius ← 6371000                    ▷ Radius of the earth in meter
 3: LatDiff ← (postionOneLat − postionTwoLat) × Π/180
 4: LngDiff ← (postionOneLng − postionTwoLng) × Π/180
                                    ▷ Convert angle from degrees to radians
 5: A ← sin(LatDiff/2) × sin(LatDiff/2) + cos(postiononelat × Π/180) ×
    cos(postiontwolat × Π/180) × sin(LngDiff/2) × sin(LngDiff/2)
 6: X ← √A                                              ▷ x-axis
 7: Y ← √(1 − A)                                        ▷ Y-axis
 8: if X > 0 or y ≠ 0 then     ▷ The angle θ between the ray to the point (x,y)
 9:     ATAN2 ← 2 * arctan(Y/√(X² + Y²) + X)
10: else if X < 0 and Y = 0 then
11:     ATAN2 ← Π
12: else if X = 0 and Y = 0 then
13:     ATAN2 ← 0
14: end if
15: C ← 2 × ATAN2
16: distance ← earthRadius × C                          ▷ Distance in meter
17: return distance
```

Fig. 4. Distance computation algorithm.

Using DL within a mobile application through the flutter framework could be very complicated. One of the challenges is focusing too much on creating algorithms and theories because the framework and libraries are new and require much more development. Another challenge is collecting a good dataset to avoid getting bad predictions and making the wrong decisions. Using DL on the edge will help the proposed mobile application to recognize holy places without connecting to Internet, which is the case for most visitors and pilgrims.

IV. EXPERIMENTAL RESULTS AND DISCUSSION

To develop the proposed mobile application, a set of tools are used such as Visual Studio Code, Flutter framework, Dart programing language, Firebase, ML kit, Automated Machine learning (Auto ML), and Genymotion emulator.

In order to validate the proposed application, a dataset describing holy places and their corresponding doaas is collected. This dataset will be used to train the DL to identify the holy places.

Firstly, we use Visual Studio Code (VSC) as a code editor, which is developed by Microsoft. VSC comes with built-in support for JavaScript, TypeScript, and Node.js. It supports other programming languages such as Dart, Python, Kotlin, and Java [27]. In addition, Flutter, a Google portable UI toolkit for building natively compiled applications for mobile, web, and desktop from a single codebase [28], is used to develop our mobile application. Flutter provides developers with full control over every pixel on the screen. It provides hot reload, which helps developers to quickly and easily experiment, add features, and fix bugs. In addition, Flutter runs 120 fps (frame per second) instead of 60fps, compared to other frameworks.

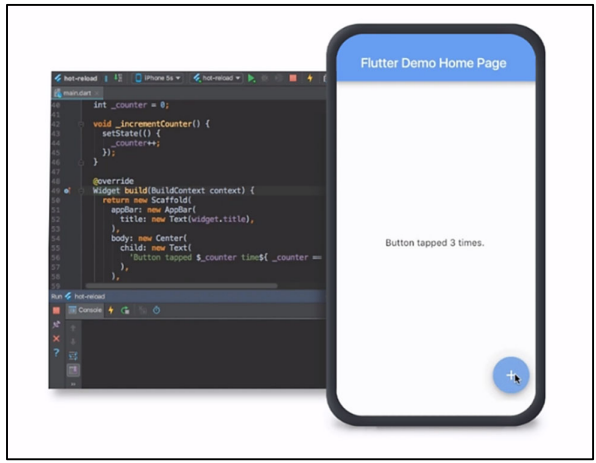

Fig. 5.  Flutter Framework.

The Dart, a client-optimized and object-oriented language for fast applications on any platform (Web, Desktop, Mobile, and Embedded) [29], is used as a programming language in our mobile application (see Fig. 6). To store doaas, we use Firebase, which is a cloud service designed for real-time and collaborative applications [30].

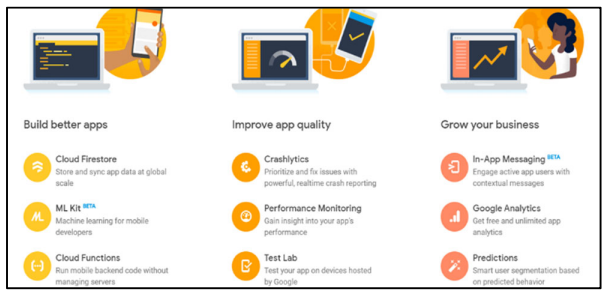

Fig. 6.  Dart Language.

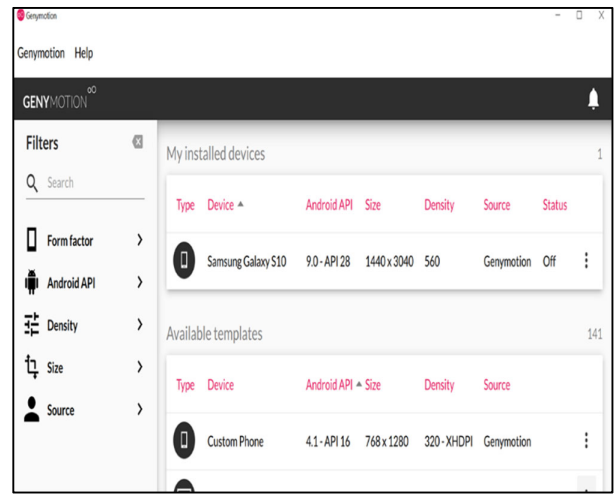

Fig. 7.  Firebase services.

Additionally, ML Kit is used as a mobile SDK that brings Google's machine learning expertise to Android and iOS apps in a powerful and ant easy-to-use manner [31]. ML Kit allows the developer to implement machine learning techniques in the application without the need to have a deep knowledge of machine learning.

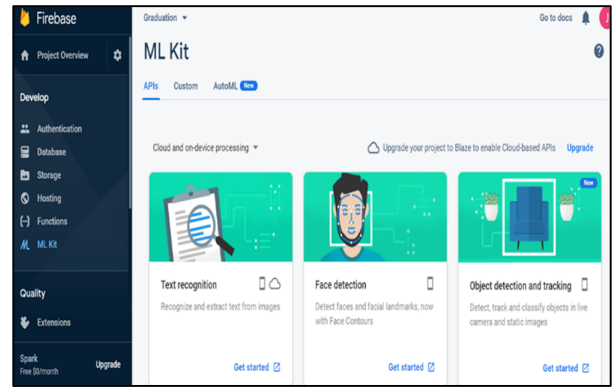

Fig. 8.  ML Kit service.

Automated Machine Learning (Auto ML) is used to allow automating the process of applying ML to real-world problems.

Finally, Genymotion is an android emulator that has been designed to help developer test their applications within a virtual environment. Genymotion provides various versions of the Android operating system. The version used in this study is 3.0.4 as depicted in Fig. 9.

Fig. 9.  Genymotion Emulator

To validate the object recognition proposed by our application, 689 images representing the holy places of Kaaba, Zamzam, and Maqam Ibrahim were collected. These images will be used during the training process. The model training process went through a number of steps as follows:

- An ML Kit project is created as depicted in Fig. 10.
- The dataset is added to the spark dataset as depicted in Fig. 11.
- Single-label classification is selected, where only one label will be assigned to the object so that the DL model can be trained to recognize this object through the label as depicted in Fig. 12.
- Import the dataset to train the DL model as in Fig. 13,
- Provide a name for each label so that the DL model is trained to recognize images by the name of the label.

- Choose a training approach among the three options as depicted in Fig. 14 depending on the specific use case. The first option is the Lowest-latency, where the accuracy is low compared to the speed of training. The General-purpose is characterized by a balance between accuracy and speed (recommended choice). Finally, the Higher-accuracy mode is used for higher prediction quality.
- Once the training process is successfully completed, the DL model is published for later use in the proposed mobile application as shown in Fig. 15 and Fig. 16.

The home screen of the proposed mobile application is presented in Fig 17.

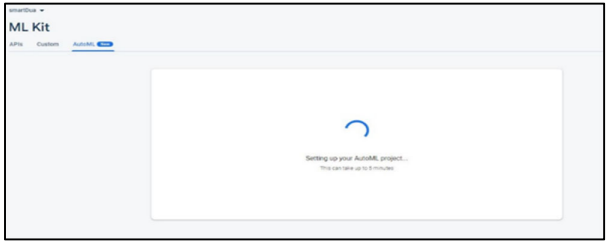

Fig. 10. Create ML Kit project

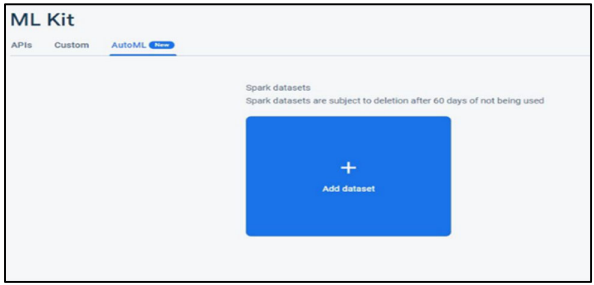

Fig. 11. Spark dataset adding

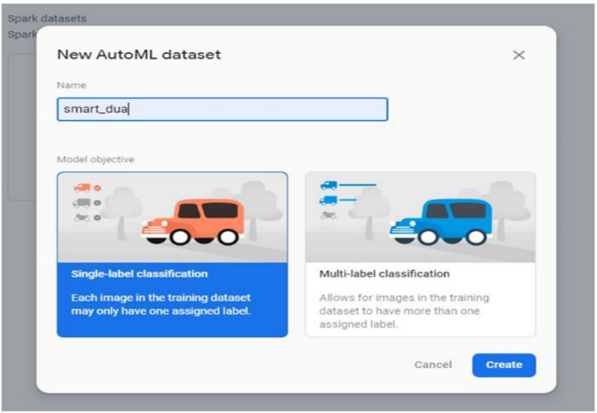

Fig. 12. Create the dataset

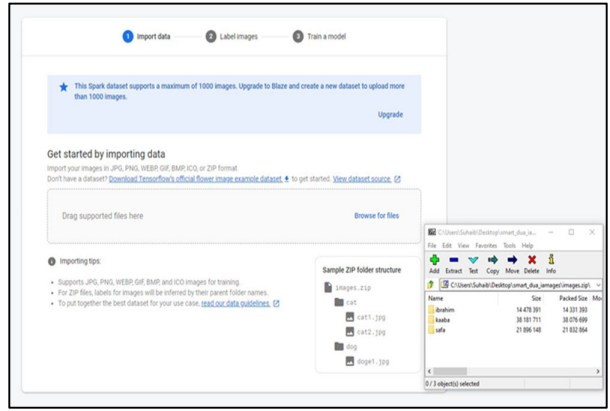

Fig. 13. Data import.

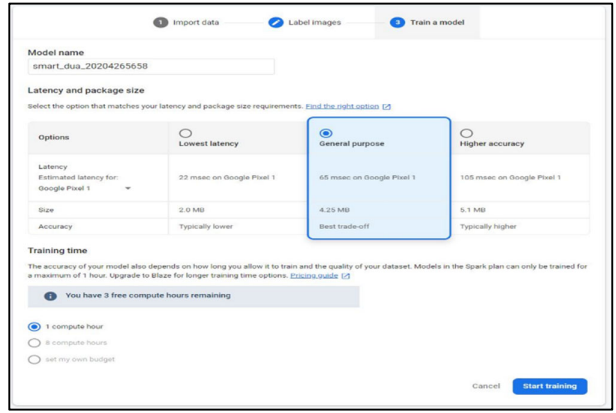

Fig. 14. Training approach.

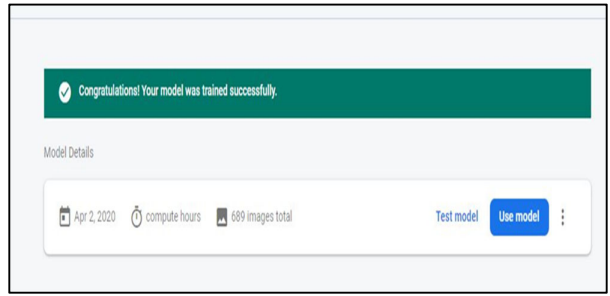

Fig. 15. Training process

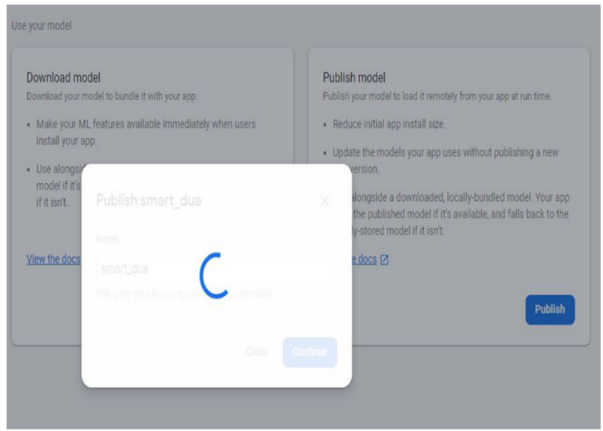

Fig. 16. Publishing of the model

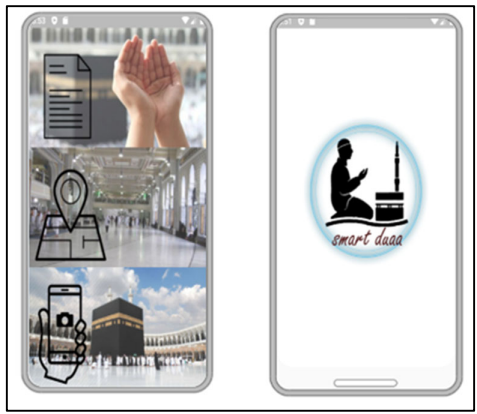

Fig. 17. Screen of the homepage

## V. CONCLUSION

In this paper, a mobile application is developed to support visitors and pilgrims in their rituals. The proposed application displays appropriate doaas corresponding to visited holy places. Three options are developed to provide appropriate doaas, which are determining the appropriate doaas by a manual search, using the GPS location to identify the holy place of the visitor and therefore its corresponding doaas, and 3) using a DL-based method to determine the holy place by analyzing an image taken by the visitor.

Results show good performances of the proposed application in determining the visited holy places by either using GPS or DL object recognition.

As future works, we plan to collect more datasets about holy places and to test the performance of the proposed mobile application in real cases. An important perspective would be to test the proposed solution in the context of displaying information about tourist places to visit.

## REFERENCES

[1] General authority for statistics, kingdom of saudi arabia, https://www.stats.gov.sa/sites/default/files/haj_40_en.pdf or https://www.stats.gov.sa/ar/indicators/196, last seen 1 February 2020.

[2] Ministry of hajj and umrah, kingdom of saudi arabia https://haj.gov.sa/ar/News/Details/12360, last seen 3 February 2020.

[3] G.W-H. Tan, V.H. Lee, B. Lin and K-B. Ooi, "Mobile applications in tourism: the future of the tourism industry?," Industrial Management & Data Systems, Industrial Management & Data Systems vol.7, no. 3, pp. 560–581, April 2017.

[4] H.Yuan, Y. Tang, W. Sun and L. Liu, "A detection method for android application security based on TF-IDF and machine learning," PLoS ONE, vol.15, no. 9, e0238694, September 2020.

[5] T. Sharma and D. Rattan, "Malicious application detection in android - A systematic literature review," Computer Science Review, vol. 40, 100373, May 2021.

[6] A. Shrestha and A. Mahmood, "Review of Deep Learning Algorithms and Architectures," IEEE Access, vol. 7, pp. 53040–53065, April 2019.

[7] J.H. alkhateeb, A.A Turani, A. Abuhamdah, M.J. Abu Sara and M.F.J. Klaib, An Effective Deep Learning Approach for Improving Off-Line Arabic Handwritten Character Recognition," International Journal Of Software Engineering & Computer Systems (IJSECS), vol. 6, no. 2, pp. 104–112, February 2021.

[8] S. Voghoei, N.H. Tonekaboni, J.G. Wallace and H.R. Arabnia, "Deep Learning at the Edge," arXiv, Computer Science-Machine Learning, October 2019.

[9] https://www.kdnuggets.com/2018/09/deep-learning-edge.html, last seen 20 February 2020.

[10] F. Waldner and F.I. Diakogiannis, "Deep learning on edge: Extracting field boundaries from satellite images with a convolutional neural network," Remote Sensing of Environment, vol. 245, pp. 111741, August 2020.

[11] J. Schmidhuber, "Deep learning in neural networks: An overview," Neural Networks, vol. 61, pp. 85–117, January 2015.

[12] H. Suk, "Chapter 1 - An Introduction to Neural Networks and Deep Learning," Deep Learning for Medical Image Analysis, Academic press, pp. 3–24, 2017.

[13] S.A. MD, "Artificial Intelligence, Machine Learning, Deep Learning, and Cognitive Computing: What Do These Terms Mean and How Will They Impact Health Care?," The Journal of Arthroplasty, vol. 33, no. 8, pp. 2358–2361, August 2018.

[14] M. Al-Sarem, W. Boulila, M. Al-Harby, J. Qadir and A. Alsaeedi, "Deep learning-based rumor detection on microblogging platforms: a systematic review," IEEE Access, vol. 7, pp. 152788-152812, 2019.

[15] S. Ben Atitallah, M. Driss, W. Boulila and H. Ben Ghézala, "Leveraging Deep Learning and IoT big data analytics to support the smart cities development: Review and future directions, " Computer Science Review, vol. 38, pp. 100303, 2020.

[16] Y. Hajjaji, W. Boulila, I. R. Farah, I. Romdhani and A. Hussain, "Big data and IoT-based applications in smart environments: A systematic review," Computer Science Review, vol. 39, pp. 100318, 2021.

[17] W. Boulila, M. Sellami, M. Driss, M. Al-Sarem, M. Safaei and F. A. Ghaleb, "RS-DCNN: A novel distributed convolutional-neural-networks based-approach for big remote-sensing image classification," Computers and Electronics in Agriculture, vol. 182, pp. 106014, 2021.

[18] W. Boulila, H. Ghandorh, M. A. Khan, F. Ahmed and J. Ahmad, "A Novel CNN-LSTM-based Approach to Predict Urban Expansion," Ecological Informatics, pp. 101325, 2021.

[19] T. Salonidis, N.. Desai and M.C. Chuah, "DeepCham: Collaborative Edge-Mediated Adaptive Deep Learning for Mobile Object Recognition," 2016 IEEE/ACM Symposium on Edge Computing (SEC), Washington, DC, USA, pp. 64–76, October 2016.

[20] C. Liu, Y. Cao, Y. Luo, G. Chen, V. Vokkarane, Y. Ma, S. Chen and P. Hou, "A New Deep Learning-based Food Recognition System for Dietary Assessment on An Edge Computing Service Infrastructure," IEEE Transactions on Services Computing, vol. 11, no. 2, pp. 249–261, March-April 2018.

[21] X. Zeng, K. Cao and M. Zhang, "MobileDeepPill: A Small-Footprint Mobile Deep Learning System for Recognizing Unconstrained Pill Images," MobiSys '17: the 15th Annual International Conference on Mobile Systems, Applications, and Services, June 2017 pp. 56–67.

[22] A. Picon, A. Alvarez-Gila, M. Seitz, A. Ortiz-Barredo, L. Echazarra, and A. Johannes, "Deep convolutional neural networks for mobile capture device-based crop disease classification in the wild," Computers and Electronics in Agriculture, vol. 161, pp. 280–290, June 2019.

[23] X. Ran, H. Chen, X. Zhu, Z. Liu and J. Chen, "DeepDecision: A Mobile Deep Learning Framework for Edge Video Analytics," IEEE INFOCOM 2018 - IEEE Conference on Computer Communications, Honolulu, HI, USA, pp. 1421-1429, April 2018.

[24] S.Y. Nikouei, Y. Chen, S. Song, R. Xu, B-Y Choi and T.R. Faughnan, "Real Time Human Detection as an Edge Service Enabled by a Lightweight CNN," arXiv, Apr 2018.

[25] T. Tan, and G. Cao, "FastVA: Deep Learning Video Analytics Through Edge Processing and NPU in Mobile," arXiv, Jan 2020.

[26] movable-type. "Calculate distance, bearing and more between Latitude/Longitude points," [ONLINE] Available at: https://www.movable-type.co.uk/scripts/latlong.html. [Accessed 11 March 2020]

[27] Visualstudio, https://code.visualstudio.com/, last seen 20 February 2020.

[28] Flutter, https://flutter.dev/ , last seen 25 February 2020.

[29] Dart, https://dart.dev/ , last seen 28 February 2020.

[30] Firebase, https://firebase.google.com/ , last seen 4 March 2020.

[31] https://firebase.google.com/docs/ml-kit , last seen 10 March 2020.